\documentclass{article}
\usepackage{spconf,amsmath,epsfig}
\usepackage{cite}
\def\degree{${}^{\circ}$}

\usepackage{multirow}
\usepackage{subfigure}
\usepackage{graphicx}
\usepackage{diagbox}

\begin{document}\sloppy

\def\x{{\mathbf x}}
\def\L{{\cal L}}

\title{Removing Rain in Videos: A Large-scale Database and A Two-stream ConvLSTM Approach}
%
\name{Tie Liu, Mai Xu$^\ast$\thanks{$^\ast$Corresponding author: Mai Xu (maixu@buaa.edu.cn). This work was supported by NSFC under Grants 61876013 and 61573037, and by the Fok Ying Tung Education Foundation under Grant 151061.} and Zulin Wang} 
\address{School of Electronic and Information Engineering, Beihang University, China \\\{liutie, maixu, wzulin\}@buaa.edu.cn}

\maketitle

\begin{abstract}
Rain removal has recently attracted increasing research attention, as it is able to enhance the visibility of rain videos. However, the existing learning based rain removal approaches for videos suffer from insufficient training data, especially when applying deep learning to remove rain.
In this paper, we establish a large-scale video database for rain removal (LasVR), which consists of 316 rain videos.
Then, we observe from our database that there exist the temporal correlation of clean content and similar patterns of rain across video frames.
According to these two observations, we propose a two-stream convolutional long- and short- term memory (ConvLSTM) approach for rain removal in videos. The first stream is composed of the subnet for rain detection, while the second stream is the subnet of rain removal that leverages the features from the rain detection subnet. Finally, the experimental results on both synthetic and real rain videos show the proposed approach performs better than other state-of-the-art approaches.
\end{abstract}
\section{Introduction}
\label{sec:intro}

Rain removal aims to separate rain streaks and then produce the clean content for images or videos. Since rain streaks hamper the visibility in images or videos captured with rain, rain removal has received increasing attention in the recent years. More importantly, the performance of many computer vision algorithms may be severely degraded due to the invisible content. Therefore, rain removal can be used in the computer vision tasks, such as object detection \cite{lowe1999object}, tracking \cite{comaniciu2003kernel} and segmentation \cite{murray1987scene}.
The past few years have witnessed some attempts for rain removal \cite{garg2004detection,zhang2006rain,liu2009pixel,kang2012automatic,kim2015video,luo2015removing,li2016rain,fu2017removing,yang2017deep,jiang2017novel,wei2017should,zhang2018density,fan2018residual,li2018video,chen2018robust,liu2018erase}.

The existing rain removal works can be generally categorized into two classes, i.e., image based and video based approaches. For images, most approaches ~\cite{kang2012automatic,luo2015removing,li2016rain,fu2017removing,zhang2018density} formulate  rain removal as a signal separation problem.
For videos, some early works were proposed to utilize the chromatic properties~\cite{liu2009pixel}, Gaussian mixture model (GMM)~\cite{bossu2011rain} and low-rank matrix completion~\cite{kim2015video} to remove rain.
Most recently, benefiting from the great success of deep learning, deep neural networks (DNNs) have been applied for rain removal of videos ~\cite{chen2018robust,liu2018erase}. For example, Liu  {\textit{et al.}}~\cite{liu2018erase} proposed a hybrid rain model to remove rain, in which the DNNs extract spatial features with temporal coherence of background.
In addition, Chen {\textit{et al.}}~\cite{chen2018robust} proposed aligning scene content at super pixel (SP) level and averaging the aligned SPs to obtain the intermediate derain output. In \cite{chen2018robust}, a convolutional neural network (CNN) is used to restore rain free details and then obtain the final output of clean content.
However, the generalization ability of these DNN-based rain removal approaches suffers from insufficient training data, as the existing video databases for rain removal are of small-scale.

\begin{figure}[t]
    \centering
    \includegraphics[width = 0.8\linewidth]{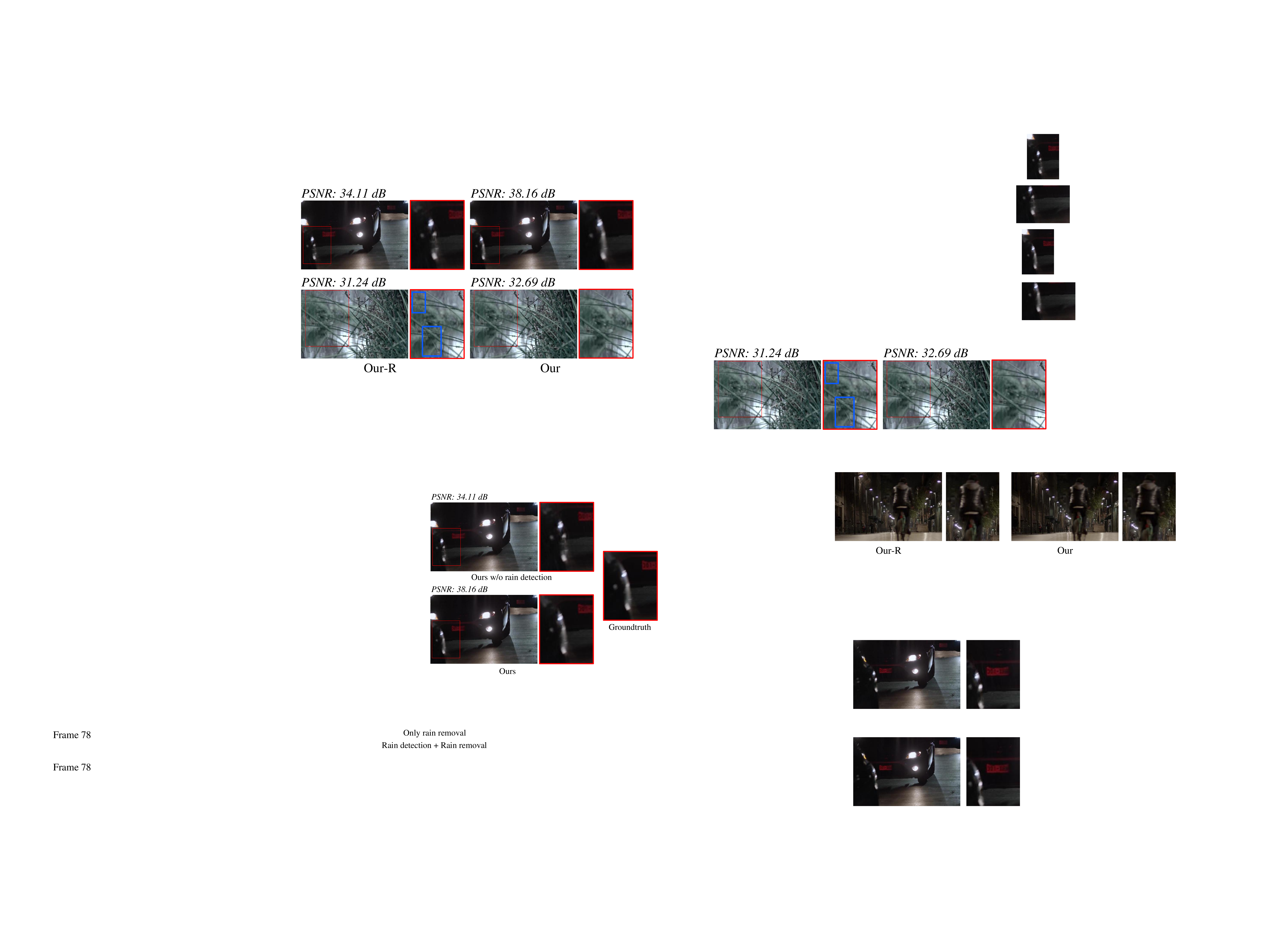}
    \vspace{-1.0em}
    \caption{\footnotesize{Comparision for visual results of our approach with and without rain detection.}}
    \label{fig:fig1_real}
    \vspace{-1.5em}
\end{figure}

To the best of our knowledge, there are only 3
video databases for rain removal,  i.e., \cite{kim2015video,chen2018robust,liu2018erase}.
Among these databases, \cite{liu2018erase} is the largest one, which has 60 videos.
In \cite{liu2018erase}, the types of rain lack diversity, since they mainly refer to \textit{direction}, \textit{scale} and \textit{density}, but not consider \textit{scene depth}, \textit{opacity}, \textit{falling speed} and \textit{wind variation}. Even worse, there exist clear horizontal boundaries in their rain videos, which is unrealistic under most circumstances.
In contrast to videos, the large-scale databases have been established for rain removal of images. For example, Fu {\textit{et al.}}~\cite{fu2017removing} created a database of 14,000 rain images. By modifying rain types, i.e., \textit{scale}, \textit{direction}, \textit{density}, \textit{scene depth}, \textit{opacity} and \textit{falling speed}, each clean image generates 14 rain images. Yang {\textit{et al.}}~\cite{yang2017deep} established a challenging database, in which each rain image contains five streak \textit{directions}. In the database of ~\cite{zhang2018density}, Zhang {\textit{et al.}} synthesized 13,200 images with three rain-density levels of rain (i.e., \textit{light}, \textit{medium} and \textit{heavy}).

\begin{figure*}
     \vspace{-2.8em}
    \centering
    \includegraphics[width = .75\linewidth]{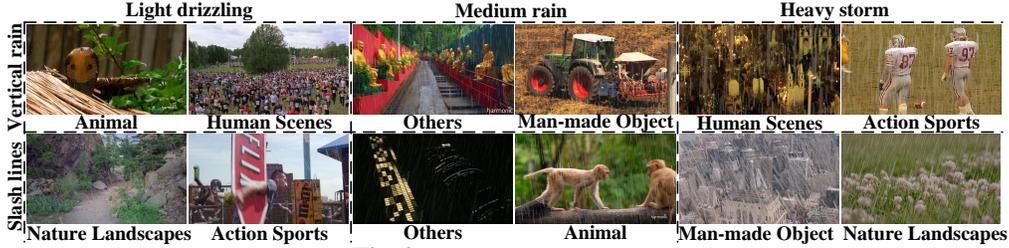}
    \vspace{-1.5em}
    \caption{\footnotesize{Examples for each category.}}
    \vspace{-0.2em}
    \label{fig:cate}
\end{figure*}

In this paper, we thus establish a large-scale database for boosting the performance of DNN-based rain removal on videos. Specifically, our database consists of 316 videos (80,835 frames) with diverse content. Moreover, various rain types are taken into account in our database, including \textit{scale}, \textit{direction}, \textit{density}, \textit{scene depth}, \textit{opacity}, \textit{falling speed} and \textit{wind variation}.
Then, we find from our database that the \textit{direction}, \textit{density} and \textit{falling speed} of rain streaks are normally invariant across video frames.
Therefore, there exist similar patterns of rain, which can be learned by convolutional LSTM (ConvLSTM).
In this paper, a two-stream ConvLSTM based approach is proposed to achieve rain detection and removal on videos.
To our best knowledge, our approach is a first attempt of applying DNNs to estimate rain streaks by leveraging the similar patterns of rain in videos.

Specifically, our approach consists of two subnets, i.e., the rain detection and rain removal subnets. The rain detection subnet leverages ConvLSTM to learn the similar patterns of rain, which can be used to estimate rain streaks in the current frame. Then, the spatial-temporal features extracted by the rain detection subnet are fused into the rain removal subnet. Consequently, the performance of rain removal can be enhanced by combining the rain detection and rain removal subnets in our approach.
Figure \ref{fig:fig1_real} illustrates the advantage of our approach in taking into account rain detection.
The extensive experiments demonstrate that our two-stream ConvLSTM approach achieves at least 4.23 dB improvement over the state-of-the-art rain removal approaches, in which 2.18 dB improvement is owing to the introduction of rain detection subnet.



The main contributions of our approach are two-fold. (1) We construct a large-scale database of 316 synthetic rain videos, from which two observations about the clean content and rain streaks are obtained. (2) We propose a novel deep learning architecture with two-stream ConvLSTM based subnets for the joint tasks of rain streak detection and removal on videos.

\begin{figure}
    \centering
    \vspace{-1.5em}
    \includegraphics[width = .7\linewidth]{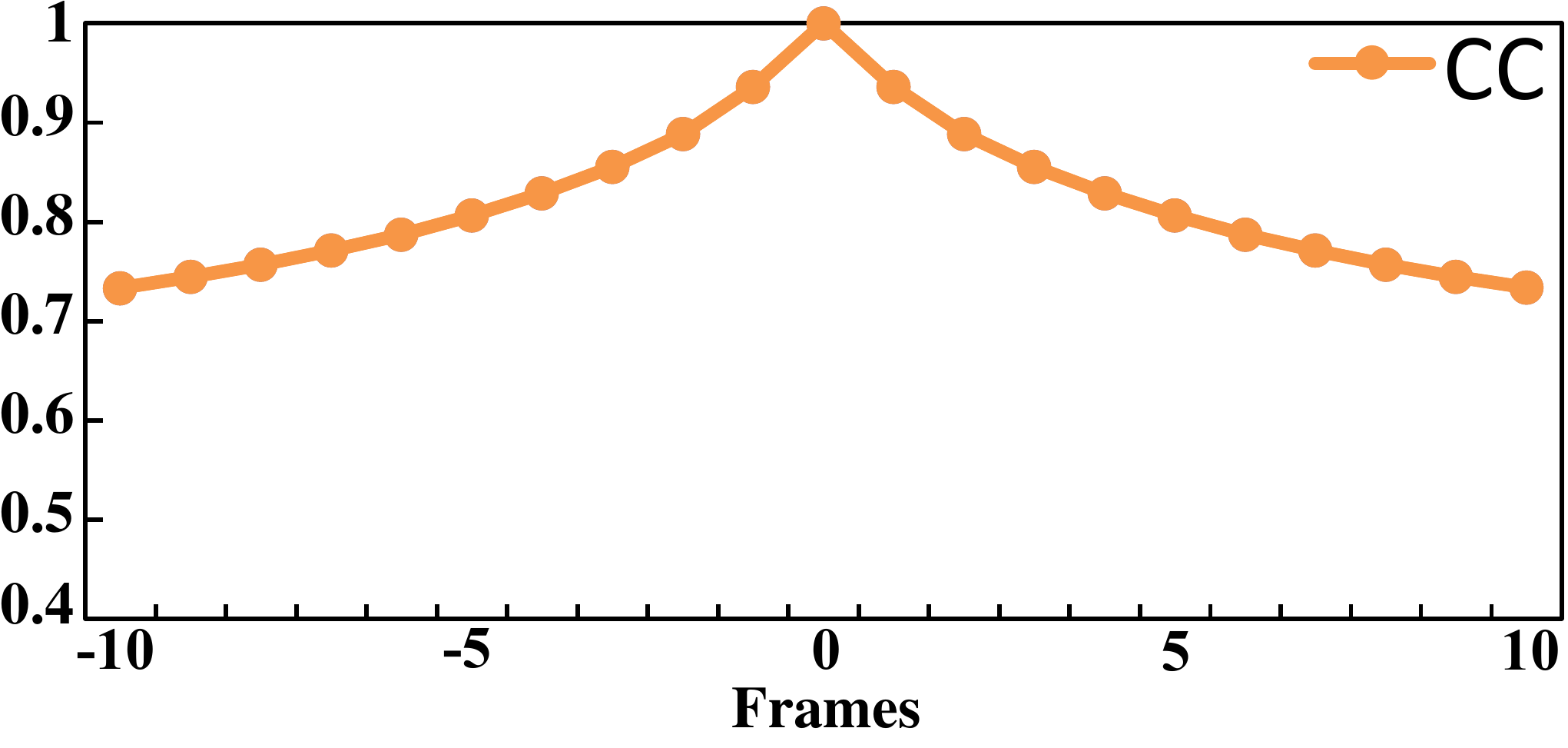}
    \vspace{-1em}
    \caption{\footnotesize{CC results of clean content across two consecutive frames with different distances (ranging from 0 to 10 frames).}}
    \vspace{-1.5em}
    \label{fig:CC}
\end{figure}

\section{LasVR Database}
The details about establishing our LasVR database are presented as follows.

\begin{figure*}
    \centering
    \vspace{-4em}
    \includegraphics[width = .99\linewidth]{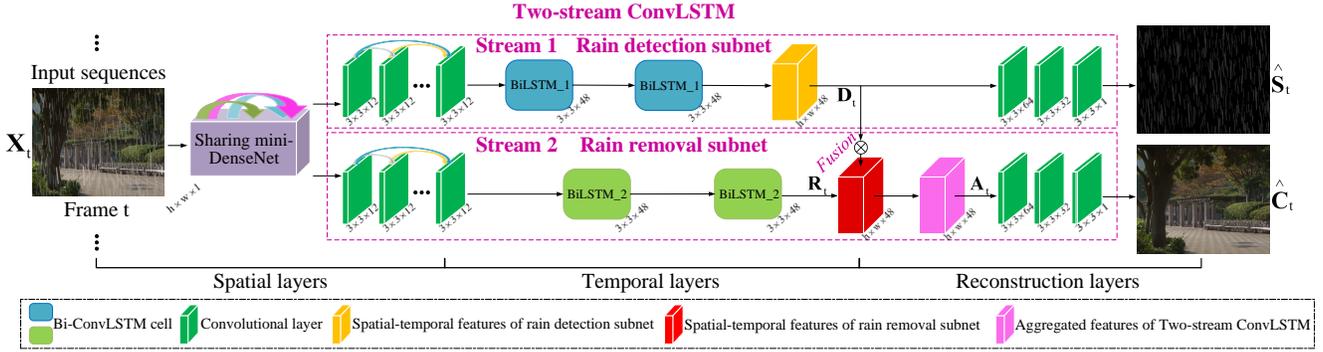}
    \vspace{-1em}
    \caption{\footnotesize{Architecture of the DNN model in our approach.}}
    \label{fig:Framework}
    \vspace{-1em}
\end{figure*}

\textbf{Clean videos.} We download 142 lossless videos from Video Quality Experts Group\footnote{https://www.its.bldrdoc.gov/vqeg/video-datasets-and-organizations.aspx/}, Harmonic\footnote{https://www.harmonicinc.com/free-4k-demo-footage/} and The Consumer Digital Video Library\footnote{https://www.cdvl.org/}. To make our database more realistic, we exclude videos with the indoor scenes. In addition, the videos of our database contain a wide range of content categories, e.g., animal, nature landscapes, human scenes, action sports, man-made object and so forth. Moreover, the resolution of our videos ranges from 640 $\times$ 360 to 2560 $\times$ 1600. In our database, videos are cut to make their duration vary from 5 to 25 seconds at the frame rate of 24-30 frames per second. Note that the bitrates are maintained above 10 Mbps when transcoding videos to the uniform MP4 format, such that the videos are with high quality.

\textbf{Rain videos.} Having obtained 142 clean videos, we randomly divide them into training (87 videos), validation (27 videos) and testing (28 videos) sets. Similar to \cite{kim2015video,chen2018robust}, we use the commercial editing software \textit{Adobe After Effect} \cite{AE} to render various types of rain for videos. To ensure the diversity of rain streaks, we set different values for the parameters of rendering, e.g., \textit{scale}, \textit{direction}, \textit{density}, \textit{scene depth}, \textit{opacity}, \textit{falling speed} and \textit{wind variation}. Consequently, the rendered streaks vary from light drizzling to heavy rain storm and vertical rain to slash line. Then, three rain videos with different parameters over each clean video are generated in the training set, while one rain video is rendered over each clean video in the validation and testing sets. Consequently, the training, validation and testing sets consist of 261, 27 and 28 rain videos. Figure \ref{fig:cate} shows some examples for each category of rain videos in our database. The peak signal-to-noise ratio (PSNR) between the rain and raw videos varies from 22.63 dB to 39.11 dB. This also implies the diversity of rain types in our database.

\section{Proposed method}
\vspace{-0.3em}
\subsection{Observations}\label{sec_observations}

We mine our database and obtain two observations on the rain videos. The observations refer to the correlation of clean video content and rain patterns, respectively.

\textbf{Temporal correlation of clean content.} It is intuitive that the clean content in consecutive frames is with high correlation. To quantify such correlation, we measure the linear correlation coefficient (CC) of the clean content between two frames. Here, the CC values between the current and the previous or subsequent frames are averaged over all clean videos in our database. Figure \ref{fig:CC} shows the CC results when the distance of two adjacent frames varies from 0 to 10.
As seen in Figure \ref{fig:CC}, the CC values are higher than 0.75 within 10 consecutive frames. This indicates the high correlation of clean content across video frames.
We further find that the temporal correlation of clean content decreases, when increasing the distance between the current and previous frames.
Therefore, there exists the long and short-term dependency of the clean content across frames. In our approach, one stream of ConvLSTM is thus used to learn such dependency for restoring the clean content of rain videos.

\textbf{Similar patterns of rain streaks.}
We further find that the rain streaks have similar characteristics for a period of time, e.g., \textit{direction}, \textit{density} and \textit{falling speed}. For \textit{density}, we measure the proportion of the rain region in a rain map by a pre-set threshold. The proportion varies from 0.1 to 0.8, while the averaged STandard Deviation (STD) at frame level is 0.04. This indicates that \textit{density} is almost the same alongside frames. For \textit{direction} and \textit{falling speed}, we select multiple Shi-Tomasi corner points from rain maps and then apply Lucas-Kanade algorithm to estimate optical flow, in order to match the positions of corner points at different time for estimating streak motion. The \textit{direction} and \textit{falling speed} range from -73\degree~to~+65\degree and 200 to 900 pixels per second, respectively. The STD results of \textit{direction} and \textit{falling speed} are 7\degree and 40 pixels per second. We can thus conclude that the \textit{direction} and \textit{falling speed} in consecutive frames remain almost invariant.

\vspace{-0.3em}
\subsection{Framework}

We assume that the frames of an input rain video are  $\mathcal{X} = \{\mathbf{X}_\textit{1}, \cdots, \mathbf{X}_t, \cdots, \mathbf{X}_\textit{T}\}$. Note that in this paper $T$ is a fixed number as the length of ConvLSTM, and each video is $T$-length segments as input.
According to the observations in Section \ref{sec_observations}, we restore the rain and clean content through two subnets (i.e., the rain detection and rain removal subnets), respectively. Assume that $\mathbf{S}_t$ is  rain streak and $\mathbf{C}_t$ is clean content at frame $t$, corresponding to $\mathbf{X}_t$. Since there exists temporal correlation across frames, we use all frames $\mathbf{X}$ to restore $\mathbf{S}_t$ and $\mathbf{C}_t$.

Specifically, the rain detection subnet integrates similar patterns of rain in consecutive frames to estimate rain streaks $\hat{\mathbf{S}}_t$ and learn spatial-temporal features of rain detection $\mathbf{D}_t$. The rain removal subnet employs temporal correlation of the clean content and spatial-temporal representation features from the other subnet to restore clean content $\hat{\mathbf{C}}_t$. Therefore, our approach jointly achieves rain detection and removal, and the combination of two stream networks significantly enhance the performance of rain removal. In this paper, the rain detection and rain removal subnets are formulated by the functions $\textit{F}_{\textbf{D}}$($\cdot$) and $\textit{F}_{\textbf{R}}$($\cdot$), respectively. Additionally, assume that $\textit{F}_{\textbf{D}}^{\textbf{S}}$($\cdot$), $\textit{F}_{\textbf{D}}^{\textbf{T}}$($\cdot$) and  $\textit{F}_{\textbf{D}}^{\textbf{R}}$($\cdot$) are the functions of spatial, temporal and reconstruction layers in the rain detection subnet, respectively. Mathematically, our approach aims at learning the following mappings:
\begin{equation}
\left\{
    \begin{aligned}
    \hat{\mathbf{S}}_t &= \textit{F}_{\textbf{D}}(\mathcal{X})= \textit{F}_{\textbf{D}}^{\textbf{R}}(\underbrace{\textit{F}_{\textbf{D}}^{\textbf{T}}(\textit{F}_{\textbf{D}}^{\textbf{S}}(\mathcal{X}))}_{\mathbf{D}_{t}})\\
    \hat{\mathbf{C}}_t &= \textit{F}_{\textbf{R}}(\mathcal{X},~\mathbf{D}_t).
    \end{aligned}
\right.
\end{equation}

\vspace{-1.3em}
\subsection{Architecture}
Figure \ref{fig:Framework} illustrates the overall architecture of our rain removal approach. Both the subnets of the rain detection and rain removal take sequences $\mathcal{X}$ as input. First, $\mathcal{X}$ is fed into a sharing mini-DenseNet for encoding the spatial features of each frame. Subsequently, the features learned by the sharing mini-DenseNet flow into two streams, corresponding to rain detection and rain removal, respectively.
In each stream, a dense unit is built to further extract spatial features for rain detection or removal. Then, the output features from the dense unit of each stream are input to bi-directional ConvLSTM. Consequently, the spatial-temporal features of rain detection and removal, denoted as ${\mathbf{D}_{t}}$ and ${\mathbf{R}_{t}}$, are obtained in two subnets. Furthermore, we fuse ${\mathbf{D}_{t}}$ and ${\mathbf{R}_{t}}$ to obtain aggregated features ${\mathbf{A}_{t}}$. Finally, ${\mathbf{D}_{t}}$ and ${\mathbf{A}_{t}}$ are used to restore rain streaks and the clean content via the reconstruction layers.

\begin{figure}
\vspace{-1em}
    \centering
    \includegraphics[width = 0.99\linewidth]{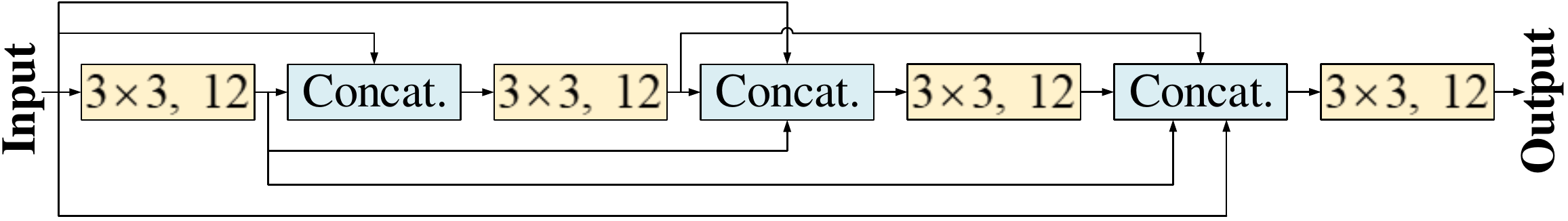}
    \vspace{-1.5em}
    \caption{\footnotesize{Structure for the dense unit.}}
    \label{fig:dense_lstm}
\vspace{-2em}
\end{figure}

\textbf{Spatial layers.} The DenseNet \cite{huang2017densely} introduces various length of inter-layer connections, for encouraging feature reuse and alleviating vanishing gradients. Considering this advantage, we apply the dense units illustrated in Figure \ref{fig:dense_lstm} to extract spatial features for rain detection and removal, respectively. The spatial layers contain the sharing mini-DenseNet and two dense units. Here, all the dense units are with the same structure. Each dense unit contains 4 convolutional layers, and features from all preceding layers are concatenated together before each layer. Hence, each dense unit consists of 10 inter-layer connections, much more than a 4-layer plain CNN with only 4 connections. At each layer in the dense unit, the number of output channels is 12. Note that the convolutional layers for each time step share the same weights and biases.

\textbf{Temporal layers.}
Section \ref{sec_observations} illustrates the temporal consistency of clean content and similar patterns of rain in consecutive frames. Hence, we adopt the ConvLSTM to learn temporal features of clean content and rain. The ConvLSTM utilizes the convolutional operation (denoted as $\ast$) in its gate computation, instead of the Hadamard product. This can preserve the spatial representation of the input frame. For the ConvLSTM layer \textit{i} at frame \textit{t}, $\mathbf{Z}_i^t$ denotes the input features; $\mathbf{I}_i^t$, $\mathbf{F}_i^t$ and $\mathbf{O}_i^t$ are the gates of input (\textit{I}), forget (\textit{F}) and output (\textit{O}); $\mathbf{G}_i^t$, $\mathbf{C}_i^t$ and $\mathbf{H}_i^t$ are the corresponding input modulation (\textit{G}), memory cell (\textit{M}) and hidden state (\textit{H}), respectively. The unidirectional ConvLSTM of layer \textit{i} at frame \textit{t} can be formulated as follows:
\begin{align}
\label{eq:convlstm_I}
\mathbf{I}_i^t &= \sigma(\mathbf{W}_i^\textit{ZI} \ast \mathbf{Z}_i^t + \mathbf{W}_i^\textit{HI} \ast \mathbf{H}_i^{t-1} + \mathbf{B}_\textit{I}),\\
\label{eq:convlstm_A}
\mathbf{F}_i^t &= \sigma(\mathbf{W}_i^\textit{ZF} \ast \mathbf{Z}_i^t + \mathbf{W}_i^\textit{HF} \ast \mathbf{H}_i^{t-1} + \mathbf{B}_\textit{F}),\\
\label{eq:convlstm_O}
\mathbf{O}_i^t &= \sigma(\mathbf{W}_i^\textit{ZO} \ast \mathbf{Z}_i^t + \mathbf{W}_i^\textit{HO} \ast \mathbf{H}_i^{t-1} + \mathbf{B}_\textit{O}),\\
\label{eq:convlstm_G}
\mathbf{G}_i^t &= {\rm tanh}(\mathbf{W}_i^\textit{ZG} \ast \mathbf{Z}_i^t + \mathbf{W}_i^\textit{HG} \ast \mathbf{H}_i^{t-1} + \mathbf{B}_\textit{G}),\\
\label{eq:convlstm_C}
\mathbf{M}_i^t &= \mathbf{F}_i^t \odot \mathbf{M}_i^{t-1} + \mathbf{I}_i^t \odot \mathbf{G}_i^t,\\
\label{eq:convlstm_H}
\mathbf{H}_i^t &= \mathbf{O}_i^t \odot {\rm tanh}(\mathbf{M}_i^t),
\end{align}
where $\odot$ denotes the element-wise multiplication. In addition, $\sigma$($\cdot$) and tanh($\cdot$) are the activation functions of sigmoid and hyperbolic tangent, respectively. Furthermore, the weights and biases of the \textit{i}-th ConvLSTM layer are denoted as $\{\mathbf{W}_i^\textit{ZI}, \mathbf{W}_i^\textit{HI}, \mathbf{W}_i^\textit{ZF}, \mathbf{W}_i^\textit{HF}, \mathbf{W}_i^\textit{ZO}, \mathbf{W}_i^\textit{HO}, \mathbf{W}_i^\textit{ZG}, \mathbf{W}_i^\textit{HG}\}$ and $\{\mathbf{B}_\textit{I}, \mathbf{B}_\textit{F}, \mathbf{B}_\textit{O}, \mathbf{B}_\textit{G}\}$.

In our approach, the ConvLSTM is built on top of the spatial layers, and each stream consists of a 2-layer ConvLSTM. The number of feature maps is 48, and the kernel size is 3 $\times$ 3. We further extend the above unidirectional ConvLSTM to work in a bidirectional fashion.

\textbf{Feature fusion.}
As discussed above, two streams of subnets aim at rain detection and removal. To combine these two streams together, we fuse the spatial-temporal features ${\mathbf{D}_{t}}$ from the rain detection subnet into the rain removal subnet. The fusion process is presented as follows,
\begin{equation}
{\mathbf{A}_{t}} = {\mathbf{R}_{t}} \odot ({\mathbf{D}_{t}} \cdot (1 - \theta) + \textbf{1} \cdot \theta),
\label{eq:feature fusion}
\end{equation}
where $\theta~(0 \leq \theta \leq$ 1) is an adjustable hyper-parameter for controlling the fusion degree, and \textbf{1} is the matrix with all elements being 1. The aggregated features ${\mathbf{A}_{t}}$ can be used for restoring clean content.

\begin{figure*}
    \centering
    \vspace{-2.5em}
    \includegraphics[width = .9\linewidth]{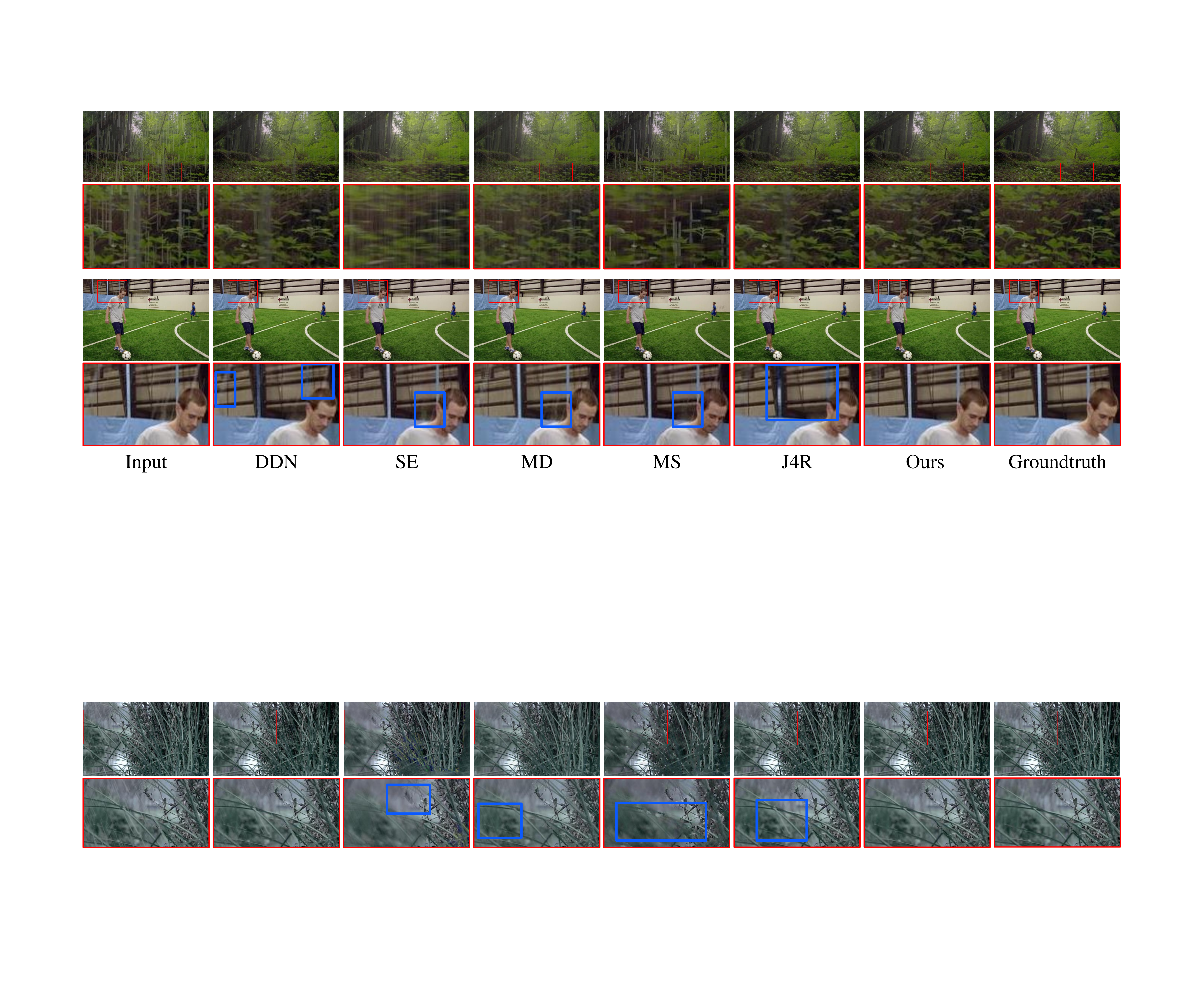}
    \vspace{-1.5em}
    \caption{\footnotesize{Comparison of different approaches on 2 example frames in our LasVR database.}}
    \label{fig:subject_syn}
\end{figure*}

\begin{table*}[t]   
\vspace{-2em}
\begin{center}
\caption{\footnotesize{Quantitative results among rain removal approaches on our database. Best results are marked in bold and the second best results are underlined.}}
\label{tab:PSNR/SSIM_syn_our}
\footnotesize
\begin{tabular}{|c|c|c|c|c|c|c|c|}
    \hline
      & Rain videos & DDN & SE & MD & MS & J4R & Ours \\
    \hline
    PSNR (dB) & 29.9404 & 36.2546 & 28.5295 & 30.9407 & 27.6112 & \underline{38.4150} & \textbf{42.6486}\\
    \hline
    SSIM & 0.8703 & 0.9528 & 0.8566 & 0.8985 & 0.7969 & \underline{0.9688} & \textbf{0.9848}\\
    \hline
\end{tabular}
\end{center}
\vspace{-2em}
\end{table*}

\textbf{Reconstruction layers.}
Finally, a K-layer CNN is exploited to estimate rain streaks and restore clean content following the temporal layers. To be more specific, the reconstruction layers in the rain removal subnet can be formulated as:
\begin{align}
&{\rm Y}_{k}(\mathbf{A}_{t}) = {\rm PReLU}(\mathbf{W}_{k} \ast {\rm Y}_{k-1}(\mathbf{A}_{t}) + \mathbf{B}_\textit{k}),1\leq k \leq K, \nonumber \\
&{\rm Y}_{K}(\mathbf{A}_{t}) = \mathbf{W}_{K} \ast {\rm Y}_{K-1}(\mathbf{A}_{t}) + \mathbf{B}_\textit{K},
\label{eq: restruction layers}
\end{align}
where the output of the \textit{k}-th convolutional layer is denoted as ${\rm Y}_{k}(\mathbf{A}_{t})$. Moreover, PReLU denotes the activation function of Parametric Rectified Linear Unit, and $\mathbf{W}_{k}$ and $\mathbf{B}_{k}$ denote the weights and biases for the \textit{k}-th convolutional layer. $K$ is set to 3 and the kernel size is 3 $\times$ 3. Consequently, the clean content at frame $t$ can be restored as $\hat{\mathbf{C}}_t = {\rm Y}_{K}(\mathbf{A}_{t})$.
Moreover, the reconstruction layers in the rain detection subnet use the same structure as that in the rain removal subnet.

\subsection{Training strategy}
In our approach, the two subnets are trained jointly in an end-to-end manner. Recall that the restored rain streaks and clean content of the current frame as $\hat{\mathbf{S}}_t$ and $\hat{\mathbf{C}}_t$, respectively. In addition, $\mathbf{S}_t$ and $\mathbf{C}_t$ are the ground-truth rain streaks and clean content, respectively. The loss function of training our DNN is denoted as:

\begin{equation}
\mathcal{L}_{t} = \alpha \cdot \underbrace{\|\hat{\mathbf{S}}_t - \mathbf{S}_t\|_{2}^{2}}_{\mathcal{L}_{\rm D}}~+~\beta \cdot \underbrace{\|\hat{\mathbf{C}}_t - \mathbf{C}_t\|_{2}^{2}}_{\mathcal{L}_{\rm R}},
\label{eq:loss}
\end{equation}
where $\alpha$ and $\beta$ are the weights of loss in restoring $\mathbf{S}_t$ and $\mathbf{C}_t$, respectively. In \eqref{eq:loss}, the loss function is calculated by the weighted sum of $\mathcal{L}_{\rm D}$ and $\mathcal{L}_{\rm R}$, which are $\ell_2$ norm error of restoring $\mathbf{S}_t$ and $\mathbf{C}_t$ by $\hat{\mathbf{S}}_t$ and $\hat{\mathbf{C}}_t$.
Since the rain detection subnet can enhance the performance of the rain removal subnet, we set $\alpha \gg \beta$ to speed up the convergence of the rain detection subnet at the beginning of training. After convergence of rain detection, we set $\alpha \ll \beta$ to minimize the $\ell_2$ norm error between $\hat{\mathbf{C}}_t$ and $\mathbf{C}_t$. Finally, the clean content can be obtained through the two subnets.

\section{Experiments}
In this section, we present the experimental results to validate the effectiveness of our two-stream ConvLSTM approach.
The experiments are conducted over our LasVR database, which is composed of 261 training videos, 27 validation videos and 28 test videos. The 261 training videos are cropped into 64 $\times$ 64 $\times$ 9 cubes, and then we have 34,800 training cubes in total.
The batch size is set to 16. We adopt the Adam optimizer with the initial learning rate as $10^{-4}$ to minimize the loss function of (\ref{eq:loss}). In the training stage, we initially set $\alpha$ = 1 and $\beta$ = 0.01 of (\ref{eq:loss}) to train the rain detection subnet. After the rain detection subnet converges, these hyperparameters are set as $\alpha$ = 0.01 and $\beta$ = 1 to train the rain removal subnet.

\textbf{Quantitative evaluation}
We evaluate the performance of our approach by comparing with 5 state-of-the-art approaches: deep detail network (DDN) \cite{fu2017removing}, stochastic encoding (SE) \cite{wei2017should}, matrix decomposition (MD) \cite{ren2017video}, multi-scale convolutional sparse coding (MS) \cite{li2018video}, and joint recurrent rain removal and reconstruction (J4R) \cite{liu2018erase}. Among them, \cite{fu2017removing} is a latest image-based approach, while others are video-based approaches. Note that we retrain the models of DDN and J4R over our database. The performance is evaluated over all test videos in our database. The evaluation metrics are PSNR and structure similarity index (SSIM). The results are reported in Table \ref{tab:PSNR/SSIM_syn_our}. As seen in Table \ref{tab:PSNR/SSIM_syn_our}, the DNN-based approaches (i.e., J4R, DDN and our approach) significantly outperform the traditional approaches (SE, MD and MS).
In addition, our approach further improves PSNR by 6.39 dB and 4.23 dB over DDN and J4R, respectively.



\begin{table}   
\vspace{-1em}
\begin{center}
\caption{\footnotesize{Quantitative results among rain removal approaches on the database \cite{chen2018robust}.}}
\label{tab:PSNR/SSIM_syn_chen}
\footnotesize
\begin{tabular}{|c|c|c|c|c|}
    \hline
    & Rain videos & DDN & J4R & Ours \\
    \hline
    PSNR (dB) & 28.7223 & 32.0166 & 32.4301 & \textbf{34.0161}\\
    \hline
    SSIM & 0.8806 & 0.9248 & 0.9381 & \textbf{0.9517}\\
    \hline
\end{tabular}
\end{center}
\vspace{-2.9em}
\end{table}

\begin{figure*}
    \centering
    \vspace{-1.5em}
    \includegraphics[width = .9\linewidth]{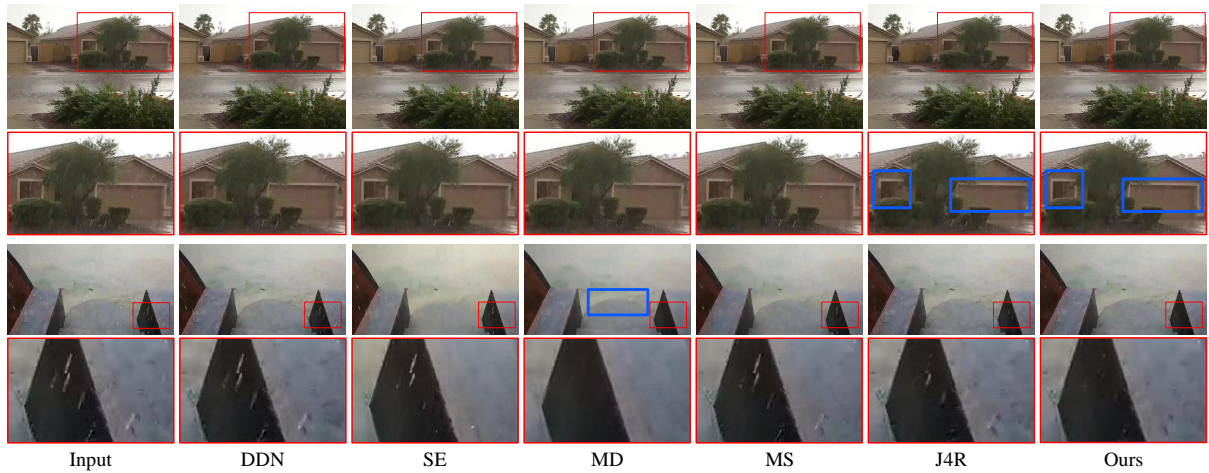}
    \vspace{-1em}
    \caption{\footnotesize{Comparison of different approaches on real videos.}}
    \label{fig:subject_real}
    \vspace{-1.5em}
\end{figure*}

\textbf{Qualitative evaluation.}
In Figure \ref{fig:subject_syn}, we present the subjective results of rain removal on 2 sample frames to visually demonstrate the results. As observed, the DNN-based approaches of J4R and DDN remove most rain streaks. However, they still fail in the removal of some rain streaks, when their opacity level is high or the structure of clean content is similar to that of rain streaks. In contrast, our approach attains promising visual effect, compared with the ground truth clean videos.
We further qualitatively evaluate the performance of different approaches on real rain videos downloaded from the Internet\footnote{https://www.youtube.com/}. As presented in Figure \ref{fig:subject_real}, MD tends to incur the blurring effect due to the evident moving objects. The SE approach still leaves some large rain streaks in frames, and the J4R approach removes most of rain streaks, but incurring detail loss. In contrast, our approach is able to remove rain streaks and meanwhile well restore the clean video frame.

\textbf{Evaluation on generalization ability.}
To evaluate the generalization ability of our approach, we further evaluate the performance of our approach and  other approaches on the database of \cite{chen2018robust}. As mentioned above, the DNN-based approaches (ours, DDN and J4R) show great advantages over the traditional approaches (MD, SE and MS), and thus we only compare our approach with DDN, J4R and SPAC-CNN \cite{chen2018robust}. As observed in Table \ref{tab:PSNR/SSIM_syn_chen}, the $\Delta$PSNR values of our approach are 2.00 dB and 1.59 dB higher than those of DDN and J4R, respectively.
In addition, the averaged $\Delta$PSNR and $\Delta$SSIM results of SPAC-CNN are 3.01 dB and 0.04 as reported in \cite{chen2018robust}, while the results of our approach are 5.29 dB and 0.07.
This demonstrates the high generalization ability of our approach for rain removal on videos.

\begin{table}
\vspace{-1em}   
\begin{center}
\caption{\footnotesize{Quantitative results compared with two baseline configurations on our database.}}
\label{tab:ablation}
\footnotesize
\begin{tabular}{|c|c|c|}
    \hline
    & Ours w/o rain detection & Ours-CNN\\
    \hline
    PSNR (dB) & 40.4702 & 41.3051\\
    \hline
    SSIM & 0.9799 & 0.9822\\
    \hline
\end{tabular}
\end{center}
\vspace{-2.5em}
\end{table}

\textbf{Ablation study.}
We test the rain removal subnet without rain detection subnet, in order to verify the effectiveness of integrating the subnet of rain detection in rain removal.
As observed in Table \ref{tab:ablation}, our approach achieves 2.18 dB higher than only rain removal in terms of PSNR. Similarly, the SSIM result of our approach is also 0.005 higher than that of only rain removal. These results indicate the performance of rain removal takes advantage of rain detection in our approach.
We further evaluate the performance of our approach replacing dense units by the plain convolutional layers (named as Ours-CNN). It can be seen that our approach increases PSNR by 1.34 dB over Ours-CNN. This verifies the effectiveness of dense units applied in our approach for rain removal in videos.

\section{Conclusion}
In this paper, we have proposed a DNN-based approach for removing rain in videos. We first established a large-scale database, which is comprised by 316 videos with diverse rain patterns, for learning the model of DNN. By investigating our database, we found two intrinsic characteristics of rain videos, i.e., content correlation and similar patterns of rain.
Then, we proposed a two-stream ConvLSTM structure that includes the rain detection and rain removal subnets. As such, rain detection and removal can be jointly achieved. More importantly, the performance of rain removal can be enhanced when detecting rain streaks and restoring clean video in a uniform DNN structure. The performance enhancement was verified through the ablation study of our experimental results. In addition, the experiments showed that our approach significantly outperforms 5  state-of-the-art approaches for both synthetic and real-world rain videos.

\footnotesize
\bibliographystyle{IEEEbib}
\bibliography{icme2019template}

\end{document}